\documentclass[10pt,twocolumn,letterpaper]{article}

\usepackage[pagenumbers]{iccv} 

\usepackage{multirow}
\definecolor{iccvblue}{rgb}{0.21,0.49,0.74}
\usepackage[pagebackref,breaklinks,colorlinks,allcolors=iccvblue]{hyperref}
\usepackage{bm}
\usepackage{multirow}
\usepackage{algorithm2e}
\usepackage{amsmath}
\usepackage{capt-of}
\usepackage{flushend, cuted}
\newcommand{\best}[1]{{{\textbf{#1}}}}
\newcommand{\second}[1]{{\underline{{#1}}}}
\def\paperID{676} 
\def\confName{ICCV}
\def\confYear{2025}
\newcommand{\NAME}{ContinuousSR\xspace}
\newcommand{\NAMECov}{DGP-Driven Covariance Weighting\xspace}
\newcommand{\NAMEPos}{Adaptive Position Drifting\xspace}
\title{Pixel to Gaussian: Ultra-Fast Continuous Super-Resolution with 2D Gaussian Modeling}

\author{Long Peng$^{1,3}$\textdagger ~~Anran Wu$^{1,2}$\textdagger ~~Wenbo Li$^{3}$\thanks{Corresponding Authors: Wenbo Li \texttt{liwenbo50@huawei.com}; Yang Wang, \texttt{ywang120@ustc.edu.cn}. \textsuperscript{\textdagger} These authors contributed equally to this work.}~~Peizhe Xia$^{1}$ ~~Xueyuan Dai$^{4}$ ~~Xinjie Zhang$^{5}$ \\ Xin Di$^{1}$ ~~Haoze Sun$^{6}$ ~~Renjing Pei$^{3}$ ~~Yang Wang$^{1,4*}$ ~~Yang Cao$^{1}$ ~~Zheng-Jun Zha$^{1}$ \\%
{$^{1}$USTC ~~$^{2}$AHU ~~$^{3}$Huawei Noah’s Ark Lab ~~$^{4}$Chang’an University~~$^{5}$HKUST~~$^{6}$THU}\\
{\small \texttt{\{longp2001@mail.,ywang120@\}ustc.edu.cn,liwenbo50@huawei.com}}\\
{\small \textbf{\url{https://github.com/peylnog/ContinuousSR}}}
}

\begin{document}
\maketitle

\begin{strip}
{\includegraphics[width=\textwidth]{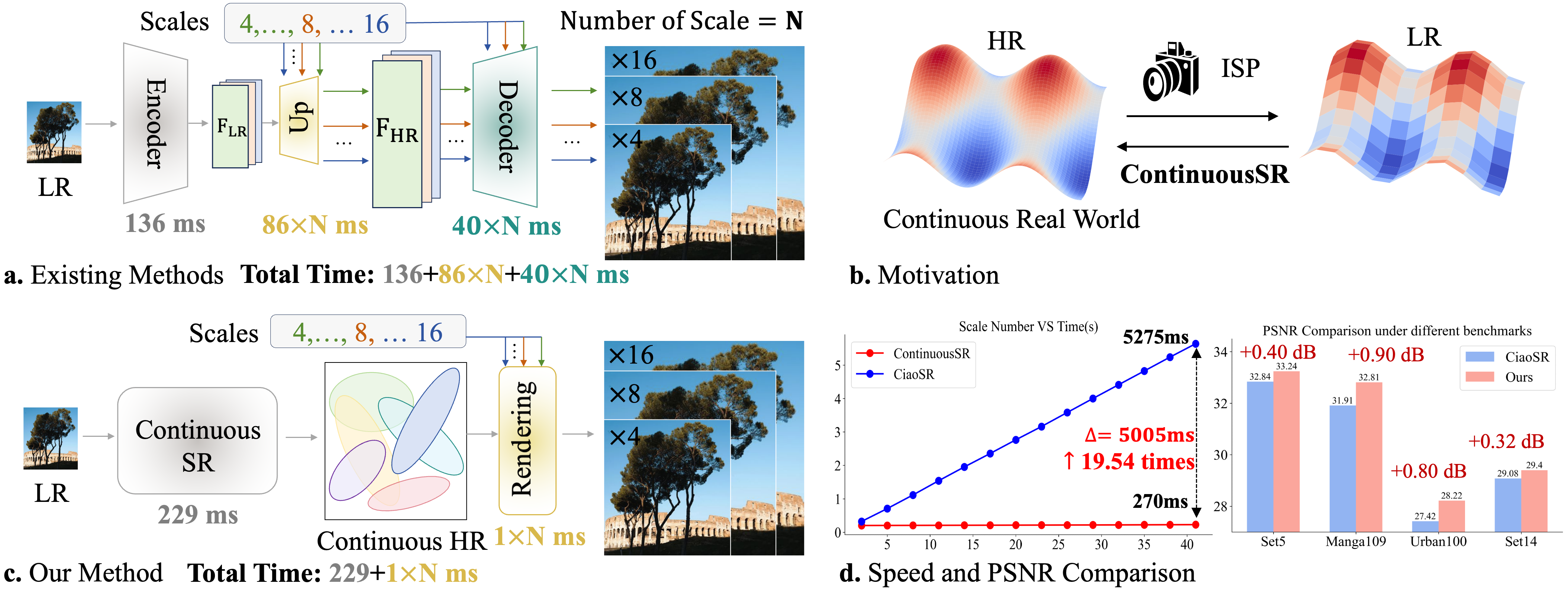}}
\vspace{-6mm}
\captionof{figure}{(a) Leveraging implicit modeling, existing ASSR methods rely on multiple upsampling and decoding steps to reconstruct HR images at different scales, which leads to low efficiency and performance. (b-d) Our method explicitly reconstructs 2D continuous HR signals from LR images in a single pass. Then, fast rendering replaces the time-consuming upsampling and decoding process to reconstruct HR images at different scales, significantly improving both performance (0.90 dB in Manga109) and efficiency (19.5× speedup).}
\label{fig:fig1}
\end{strip}

\begin{abstract}

Arbitrary-scale super-resolution (ASSR) aims to reconstruct high-resolution (HR) images from low-resolution (LR) inputs with arbitrary upsampling factors using a single model, addressing the limitations of traditional SR methods constrained to fixed-scale factors (\textit{e.g.}, $\times$ 2). Recent advances leveraging implicit neural representation (INR) have achieved great progress by modeling coordinate-to-pixel mappings. However, the efficiency of these methods may suffer from repeated upsampling and decoding, while their reconstruction fidelity and quality are constrained by the intrinsic representational limitations of coordinate-based functions. To address these challenges, we propose a novel ContinuousSR framework with a Pixel-to-Gaussian paradigm, which explicitly reconstructs 2D continuous HR signals from LR images using Gaussian Splatting. This approach eliminates the need for time-consuming upsampling and decoding, enabling extremely fast arbitrary-scale super-resolution. Once the Gaussian field is built in a single pass, ContinuousSR can perform arbitrary-scale rendering in just 1ms per scale. Our method introduces several key innovations. Through statistical analysis, we uncover the Deep Gaussian Prior (DGP) and propose DGP-Driven Covariance Weighting, which dynamically optimizes covariance via adaptive weighting. Additionally, we present Adaptive Position Drifting, which refines the positional distribution of the Gaussian space based on image content, further enhancing reconstruction quality. Extensive experiments on seven benchmarks demonstrate that our ContinuousSR delivers significant improvements in SR quality across all scales, with an impressive 19.5× speedup when continuously upsampling an image across forty scales.
\end{abstract}

\section{Introdution}

Cameras and smartphones discretize continuous real-world scenes into discrete 2D digital images~\cite{castleman1979digital,jain1989fundamentals,xia2024text}, as illustrated in Figure~\ref{fig:fig1}(b). However, limitations in sensor resolution, among other factors, often lead to low-resolution (LR) images that fail to meet user requirements. Image super-resolution (SR) has been proposed to enhance image resolution and finer details~\cite{wang2020deep,liu2022blind}. Unlike traditional fixed-scale super-resolution ~\cite{peng2024efficient,liang2021swinir,chen2023hat,lim2017edsr,zhang2018rcan}, which uses multiple models to learn mappings for fixed scales (\textit{e.g.}, $\times 2$, $\times 3$, $\times 4$), arbitrary-scale super-resolution (ASSR) employs a single model to handle super-resolution with arbitrary scales, which has attracted significant attention~\cite{hu2019metasr,wei2023super,chen2023cascaded,LTE,liu2024arbitrary,wan2024arbitrary,li2024savsr}.

Among these approaches, implicit neural representation (INR) has emerged as a leading technique, delivering visually compelling results~\cite{chen2021liif,lee2022lte,xu2021ultrasr,cao2023ciaosr,xia2024text}. INR aims to learn a continuous mapping from pixel coordinates to pixel values, enabling arbitrary-scale super-resolution through multiple upsampling and decoding steps, as illustrated in Figure~\ref{fig:fig1}(a). For example, LIIF~\cite{chen2021liif} is the first to introduce INR into ASSR, employing multi-layer perceptrons to learn this mapping. Later, CiaoSR~\cite{cao2023ciaosr} and CLIT~\cite{chen2023cascaded} leverage Transformers to enhance the modeling of long-range dependencies in feature upsampling and decoding, achieving state-of-the-art performance. However, the reconstruction fidelity and quality of these methods are inherently constrained by the representational limitations of coordinate-based implicit functions, making it challenging to effectively model continuous high-resolution signals, ultimately leading to suboptimal performance. Additionally, their reliance on repeated upsampling and decoding significantly reduces efficiency, making real-world deployment impractical.

Given that LR images are discretized from continuous 2D signals, we pose the fundamental question: ``\textit{Can we directly reconstruct continuous HR signals from LR images and flexibly choose the desired scale?}'' As illustrated in Figure~\ref{fig:fig1}(b), this approach not only enhances signal continuity through continuous modeling—leading to improved reconstruction quality—but also significantly boosts efficiency by eliminating the need for time-consuming upsampling and decoding. This idea enables fast and flexible ASSR, making real-world applications more practical.

In this paper, we introduce the novel {\NAME} framework, built upon the Pixel-to-Gaussian paradigm, which reconstructs 2D continuous HR signals through Gaussian modeling. By first reconstructing a continuous HR Gaussian field, our method enables rapid sampling directly from the continuous representation, effectively replacing traditional time-consuming upsampling and decoding steps. This innovative approach achieves high-quality ASSR in just 1 ms, significantly improving the efficiency.

Directly applying Gaussian modeling to simulate real-world images is highly challenging due to the intricate interweaving of pixel distributions and parameters. To address this, we first identify the Deep Gaussian Prior (DGP) from 40,000 natural images, revealing that the distribution of Gaussian field parameters follows a Gaussian pattern with regularities in their range, as illustrated in Figure~\ref{fig:fig2}(a-b). Leveraging this insight, we sample pre-defined Gaussian kernels from the DGP distribution and introduce a novel \NAMECov module, which efficiently optimizes covariance parameters through adaptive weighting. This helps guide the model toward the global optimum. Furthermore, we propose a \NAMEPos module, which dynamically adjusts the spatial positions of Gaussian kernels based on image content, enhancing structural accuracy. With these innovations, our method not only surpasses state-of-the-art approaches by up to 0.9 dB in reconstruction performance but also achieves a 19.5× speedup when continuously upsampling across forty scales. Our main contributions are as follows:



\begin{itemize}[leftmargin=3mm]

\item {A novel {{\NAME}} is proposed to reconstruct continuous HR signals from LR images by 2D Gaussian modeling, thereby enabling fast and high-quality super-resolution with arbitrary scale.}

\item {The Deep Gaussian Prior (DGP) is discovered, based on which \NAMECov is proposed to facilitate the optimization of covariance. Furthermore, \NAMEPos is introduced to dynamically learn spatial positions in Gaussian space.}

\item {Extensive experiments demonstrate that our method achieves state-of-the-art performance on seven benchmarks and ultra-fast speed.}
    
\end{itemize}

\section{Related work}
\subsection{{Arbitrary-Scale Super-Resolution}}

Although traditional fixed-scale super-resolution (FSSR) methods, which use separate models to learn different super-resolution scales, have achieved significant progress~\cite{dong2014srcnn,ledig2017srresnet,zhang2018rcan,kim2016vdsr,cavigelli2017cas,zhang2021DPIR,wang2018esrgan,di2024qmambabsr,peng2024towards,peng2025directing,peng2024lightweight,peng2024unveiling,wang2023decoupling,peng2020cumulative,peng2021ensemble,yi2021efficient,yi2021structure,zhang2024scrutinize,zhang2023fine,li2023ntire,ren2024ninth}, they struggle to meet the demand for arbitrary-scale super-resolution in real-world scenarios. Additionally, maintaining multiple models incurs high computational costs, making them less practical. To address these limitations, Arbitrary-Scale Super-Resolution (ASSR) has been proposed to achieve it with a single model, gaining increasing attention in recent years~\cite{cao2023ciaosr,hu2019metasr,fu2024continuous,he2024latent,zhu2025multi,zhao2024activating,tsai2024arbitrary,fu2024continuous,zhang2024deep,shang2024arbitrary,jiang2024sqformer,duan2024local,he2024dynamic}. For example, MetaSR~\cite{hu2019metasr} was the first to introduce the meta-upscale module to achieve arbitrary-scale super-resolution, demonstrating promising results. Inspired by the success of implicit neural representation (INR) in 3D reconstruction, LIIF~\cite{chen2021liif} was the first to adapt INR to super-resolution by using a multilayer perceptron to learn the mapping from image coordinates and features to RGB values. To capture more high-frequency details, LTE~\cite{lee2022lte} encodes textures in the Fourier space, while SRNO~\cite{wei2023super} leverages neural operators to model global relationships. CLIT~\cite{chen2023cascaded} introduces a cross-scale interaction mechanism to enhance feature learning by integrating information across different resolutions. CiaoSR~\cite{cao2023ciaosr} further improves long-range modeling capability by introducing transformers to INR, achieving state-of-the-art performance. LMF~\cite{he2024latent} enhances local texture details by combining multi-frequency information in a computationally efficient manner, significantly reducing computational costs while maintaining the reconstruction of fine-grained features. However, these implicit modeling methods struggle to explicitly reconstruct continuous HR signals and require time-consuming upsampling and decoding, leading to low performance and efficiency.

\begin{figure*}[ht]
  \centering
  \includegraphics[width=1\linewidth]{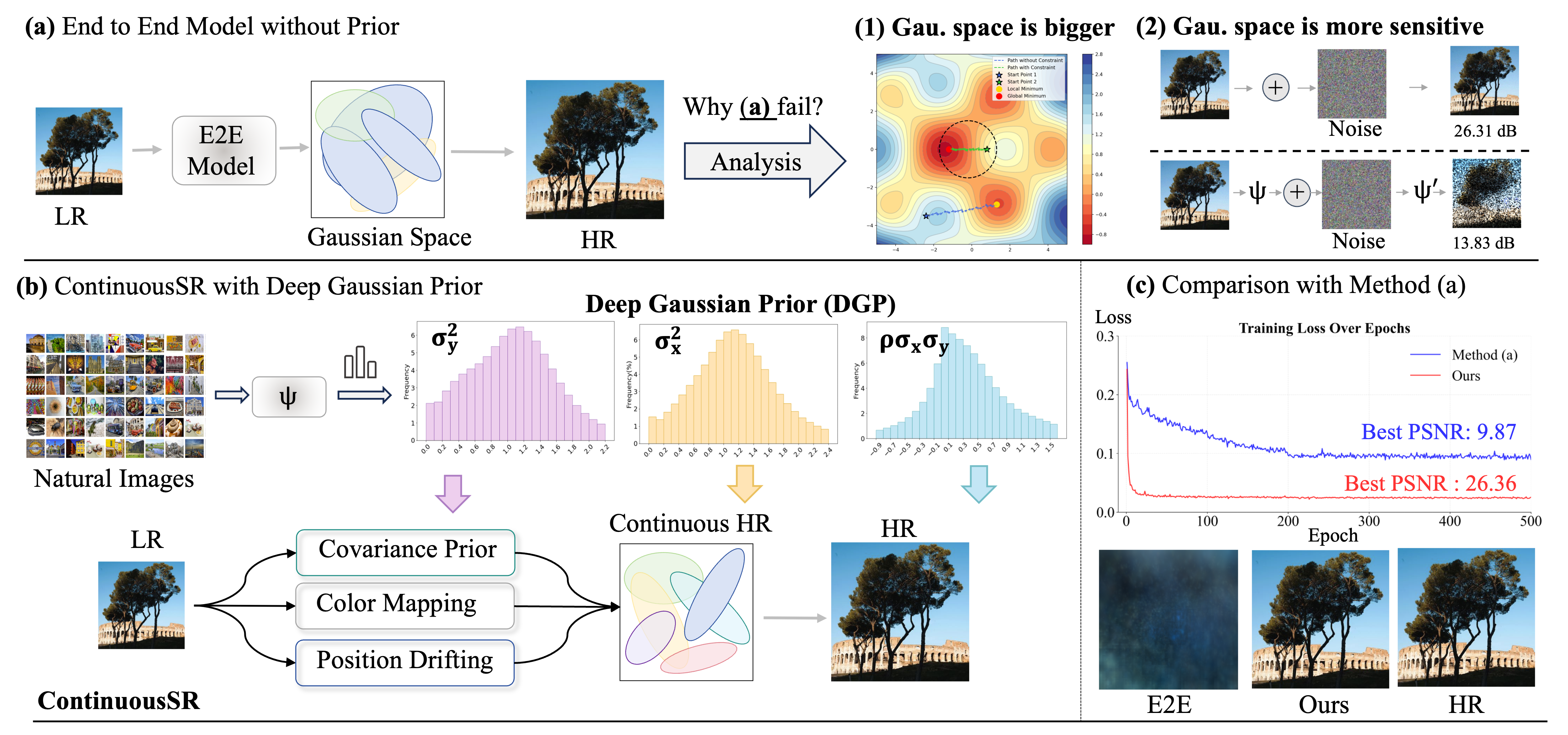}
   \vspace{-4mm}
\caption{(a) Directly learning the end-to-end model from LR to the Gaussian field is challenging due to the vastness and sensitivity of the Gaussian space. (b-c) Through statistical analysis of 40,000 natural images, we uncover the Deep Gaussian Prior and propose Position Drifting, Covariance Prior, and Color Mapping to propose a novel ContinuousSR, enhancing the quality of the Gaussian field.}
\vspace{-4mm}
  \label{fig:fig2}
\end{figure*}

\subsection{Gaussian Splatting}
Gaussian Splatting (GS) is introduced into 3D as a faster, more efficient alternative to NeRF, using anisotropic 3D Gaussians for real-time rendering and direct scene manipulation~\cite{kerbl20233d}. Building on 3DGS, 2D Gaussian Splatting improves the geometric accuracy of radiance fields by combining 2D Gaussians with precise scene projections~\cite{huang20242d}. Recently, 2D GS finds applications in image processing~\cite{chen2025generalized,dong2025gaussiantoken}. For instance, Zhang \textit{et al.} propose leveraging Gaussian Splatting (GS) for image compression and reconstruction~\cite{zhang2024gaussianimage} through long-time optimization of GS parameters, while Hu \textit{et al.} employ Gaussian Splatting in the feature space to enhance visual quality and speed~\cite{GaussianSR}. However, these methods still struggle to reconstruct continuous HR signals and suffer from long optimization times or multiple upsampling and decoding process.

\section{{Motivation}}
To capture the real world, advanced imaging sensors (\textit{e.g.}, CMOS) are used to project the 3D continuous world into 2D and then discretized 2D continuous signals into 2D discrete signals~\cite{castleman1979digital,jain1989fundamentals,xia2024text}, as formulated:
\begin{equation}
\label{eq:imaging}
I\left[ m, n \right] = f_{c}(m \Delta x, n \Delta y).
\end{equation}
where \(f_c(x, y)\) represents the continuous intensity function in the spatial domain \((x, y)\). The \(\Delta x\) and \(\Delta y\) denote the sampling step along the spatial dimensions, while \(m, n \in \mathbb{Z}\) are the theresponding discrete pixel grids. \(I\left[ m, n \right]\) represent the discrete images. After that, the Image Signal Processor is used to quantize, process, and encode it into a digital low-resolution image $\mathbf{I}_{\text{LR}}$.

Although many methods leveraging implicit modeling have been proposed~\cite{chen2021liif,LTE} to achieve ASSR by constructing coordinate-to-pixel mappings, two major challenges remain. On the one hand, the aim of ASSR is to reconstruct \(f_c(x, y)\). However, implicit modeling makes it difficult to explicitly model high-quality continuous functions, resulting in limited performance. On the other hand, the pipeline of INR-based ASSR methods suffers from low efficiency, as follows:
\begin{equation}
\begin{aligned}
\mathcal{F}_{LR} &= \mathbb{E}(\mathbf{I}_{\text{LR}}), \quad \mathcal{F}^{s}_{HR} = \mathbb{U}(\mathcal{F}_{LR}, s), \\
\mathbf{I}^{s}_{\text{HR}} &= \mathbb{D}(F^{s}_{HR}).
\end{aligned}
\end{equation}
where \(\mathbb{E}\), \(\mathbb{U}\) and \(\mathbb{D}\) represent the Encoder, Upsampling, Decoder, respectively, and \(\mathcal{F}^{s}_{HR}\) denotes the high-resolution feature map at scale \(s\). It can be observed that for different scales \(s\), this method requires multiple time-consuming upsampling \(\mathbb{U}\) and decoding \(\mathbb{D}\) processes to reconstruct HR images $F^{s}_{HR}$ , as shown in Figure~\ref{fig:fig1}(b), resulting in inefficiency. Therefore, we propose the fundamental question: ``\textit{Can we directly reconstruct continuous HR signals from LR images?}'' This serves as the inverse function of imaging process Eq.~\ref{eq:imaging}, as illustrated in Figure~\ref{fig:fig1}(c). This approach would not only perform simple sampling to replace multiple upsampling and decoding but also enhance continuity, improving efficiency and performance.

\section{{Proposed Method}}

\subsection{Continuous Basis Function}
Considering that the target function is continuous, it is crucial to select an appropriate continuous basis function. In this work, we choose the Gaussian function for two main reasons: a) Leveraging the Gaussian Mixture Model (GMM)~\cite{reynolds2009gaussian}, any complex continuous function can be represented as a combination of several Gaussian functions, ensuring broad applicability and theoretical soundness. b) With the recent advancements in the Gaussian splatting community~\cite{fei20243d,chen2024survey}, the engineering efficiency and compatibility of Gaussian functions have significantly improved, making them highly suitable for practical implementation. Therefore, we use Gaussian functions $G_i(x, y)$ as fundamental continuous functions to reconstruct real 2D continuous signals $f_{c}(x, y)$, as shown in the following equation:
\begin{equation}
f_{c}(x, y) = \sum_{i=1}^N  G_i(x,y)
\end{equation}
where $N$ denotes the number of Gaussian kernels, $x$ and $y$ represents the location in the 2D space. Each Gaussian kernel has eight parameters needed to optimized, which include:
\begin{equation}
\Sigma = \begin{bmatrix}
\sigma_{x}^2 & \rho \sigma_{x} \sigma_{y} \\
\rho \sigma_{x} \sigma_{y} & \sigma_{y}^2
\end{bmatrix}, 
\mu = \begin{bmatrix}
\mu_{x} \\
\mu_{y}
\end{bmatrix},
c_{rgb} = \begin{bmatrix}
c_{r} \\
c_{g} \\
c_{b}
\end{bmatrix},
\end{equation}
where $c_{rgb}$ denotes the RGB parameters of each Gaussian, $\mu$ represents the position parameters, and $\Sigma$ represents the covariance matrix, resulting in a total of eight parameters to be optimized. The value of the Gaussian kernel $G_i$ at the position $(x,y)$ can be expressed as:
\begin{equation}
G_i(x, y, c_{rgb},\Sigma) = c_{rgb}\frac{1}{2\pi |\Sigma_i|} \exp\left(-\frac{1}{2} 
d^\top 
\Sigma_i^{-1}
d
\right).
\end{equation}
where the distance vector $d$ represents the deviations of $x$ and $y$ from their positions $\mu_x$ and $\mu_y$.

\subsection{Direct End-to-End and Deep Gaussian Prior}
\label{sec:fail_solution}
A straightforward approach is to learn the parameters of Gaussian kernels directly from low-resolution (LR) images through an end-to-end model. However, this approach is extremely difficult to optimize, as shown in Figure~\ref{fig:fig2}(a). As shown in Figure~\ref{fig:fig2}(c), the blue loss curve indicates that the optimization process falls into a local optimum, with the PSNR remaining as low as 10 dB. To rule out the possibility of coincidence, we conduct multiple experiments and consistently observe the same conclusion.\begin{figure*}[t]
  \centering
  \vspace{-3mm}
  \includegraphics[width=1\linewidth]{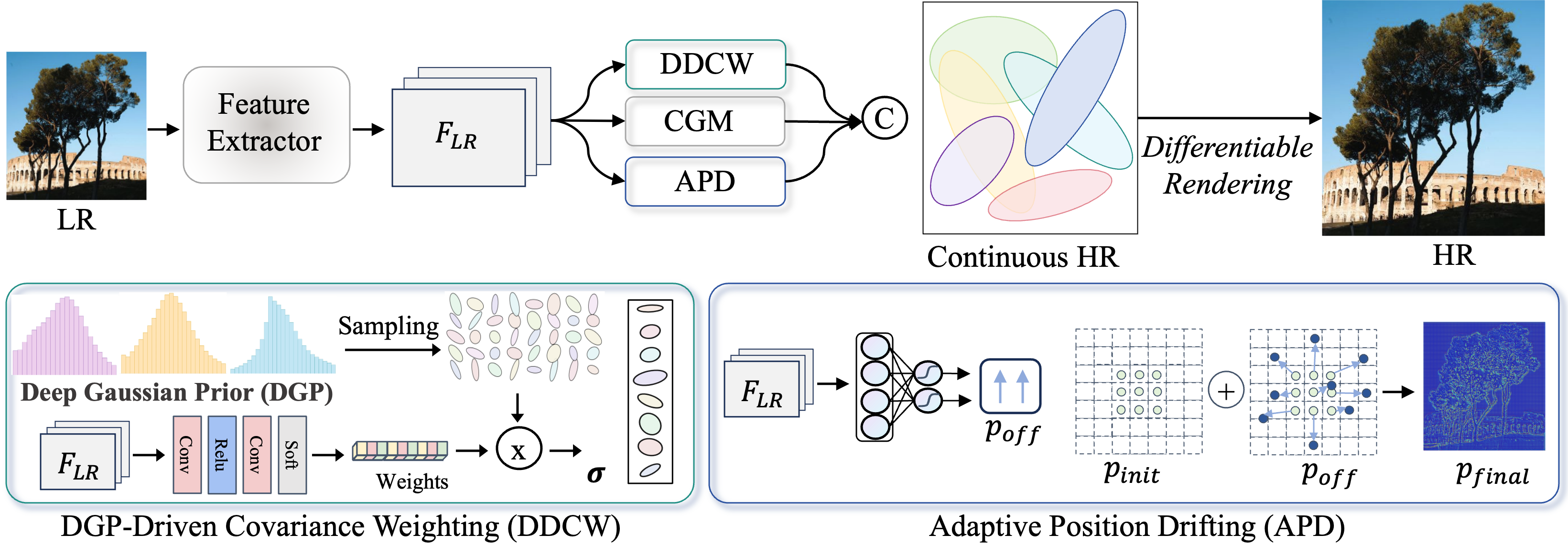}
  \vspace{-3mm}
\caption{An overview of the proposed ContinuousSR framework, which consists of three key innovations: DGP-Driven Covariance Weighting (DDCW), Adaptive Position Drifting (APD), and Color Gaussian Mapping (CGM).}
\vspace{-3mm}
  \label{fig:fig3}
\end{figure*}\\
{\bf Why does direct end-to-end fail?} We attribute this to two main challenges: \textbf{a) High Complexity}: Each kernel in the Gaussian space contains numerous difficult-to-learn parameters that need to be optimized, such as position, covariance, and RGB values.  Many of these parameters have solution spaces ranging from 0 to positive infinity, resulting in an exceptionally large solution space. For instance, the covariance matrix theoretically only needs to satisfy the condition of being a positive definite matrix. This makes the Gaussian Space significantly larger than traditional image space, while introducing more local traps. Consequently, the complexity of optimization in Gaussian Space increases, making it more prone to local optima, as illustrated in Figure~\ref{fig:fig2} (a). \textbf{b) High Sensitivity}: In Gaussian Space, even a slight change in any parameter of a single Gaussian, such as position or covariance, can significantly affect the entire image. This is fundamentally different from the image space, where a single pixel only impacts itself. To further verify this, we add noise with the same distribution to both image space and Gaussian Space to evaluate sensitivity. Note that the Gaussian Space is derived through the optimization method~\cite{zhang2024gaussianimage}, denoted $\psi$, which requires approximately 1 minute of GPU time per scene to ensure high-quality mapping. The comparison results, as shown in Figure~\ref{fig:fig2}(c), indicate that the PSNR in image space is 26.31 dB, whereas it is only 13.83 dB in Gaussian Space. This demonstrates that Gaussian Space is much more sensitive, making the optimization more challenging.\\
{\bf Observation and Deep Gaussian Prior.} To uncover the secrets of the Gaussian Space, we conduct statistical experiments to analyze the distribution of Gaussian parameters. Specifically, we collect and crop approximately 40,000 high-resolution images~\cite{timofte2017ntire,lim2017enhanced}, and transform them into the Gaussian space using $\psi$, with optimized over 700 GPU-hours. Subsequently, we statistically analyze the key parameters of Gaussian kernels, including $\sigma_x^2$, $\sigma_y^2$, and $\rho \sigma_x \sigma_y$. The results, as shown in Figure~\ref{fig:fig2}(b), indicate that the distribution of most covariances is traceable: a) Approximately 99\% of $\sigma_x^2$, $\sigma_y^2$, and $\rho \sigma_x \sigma_y$ fall within the ranges of $0\sim2.4$, $0\sim2.2$, and $-0.9\sim1.5$, respectively. b) The distributions of the three covariances generally follow a Gaussian distribution.  We define this finding as the {\bf Deep Gaussian Prior (DGP)}, which provides valuable information to reduce the difficulty of optimization. Based on these observations, we propose an innovative method, \NAME, which for the first time achieves representation learning from low-resolution (LR) images to continuous HR signals. Specifically, \NAME introduces \NAMECov, which simplifies the optimization difficulty in Gaussian Space by constructing pre-defined Gaussian kernels, employing an adaptive weighting mechanism, and incorporating \NAMEPos based on offset drifting. This approach enables superior performance and achieves fast super-resolution results, as shown in Figure~\ref{fig:fig1} and ~\ref{fig:fig3}.

\begin{table*}[t]
\centering
\vspace{-2mm}
\caption{PSNR performance comparison with state-of-the-art methods under different benchmarks. Average Time (AT) is reported in milliseconds (ms). The best and the second-best results are in \best{bold} and \second{bold}. More comparisons are in {Appendix Section~\ref{sec:Addtional_Comparison}}.}
\vspace{-2mm}
\scalebox{0.90}{
\begin{tabular}{l|l|cccccccccc|r}
\toprule
PSNR$\uparrow$ & Methods    & $\times$4    & $\times$6    & $\times$8    & $\times$10   & $\times$12   & $\times$16   & $\times$18   & $\times$20   & $\times$32   & $\times$48   & {AT} \\ \midrule
                             & MetaSR~\cite{hu2019metasr}     &  26.76     &    24.31   &   22.92    &    22.02   &  21.31     & 20.35      &   19.96    &    19.65   &  18.38     &   17.48    & \second{41.4}     \\
                             & LIIF~\cite{chen2021liif}       & 26.68 & 24.20  & 22.79 & 21.84 & 21.15 & 20.19 & 19.80  & 19.51 & 18.30  & 17.45 &  110.0                             \\
                             
                             & LTE~\cite{lee2022lte}        & 27.24 & 24.62 & 23.17 & 22.23 & 21.50  & 20.47 & 20.06 & 19.77 & \second{18.47} & \second{17.52} &   151.8                           \\
                             & SRNO~\cite{wei2023super}        &      26.98	&   24.43	&   23.02	&   22.06	&   21.36	&   20.35	&  19.95	&  19.67	&  18.39	&  17.51      &   65.7                            \\
                             & CiaoSR~\cite{cao2023ciaosr}     & \second{27.42} & \second{24.84} & \second{23.34} & \second{22.34} & \second{21.60}  & \second{20.54} & \second{20.11} & \second{19.77} & {18.45} & 17.51 &   341.5                           \\
                             & MambaSR~\cite{yan2024mambasr} &27.02 &24.44 &23.01 &22.06 &21.36 &20.34 &19.95 &19.65 &18.29 &17.48 &90.5
                             \\
                             & GaussianSR~\cite{GaussianSR} & 26.20  & 23.76 & 22.35 & 21.38 & 20.66 & 19.68 & 19.31 & 19.03 & 17.86 & 17.07 & 321.4                              \\
\multirow{-8}{*}{Urban100~\cite{huang2015single}}   & Ours       & \best{28.22}	& \best{25.43} & \best{23.87} & \best{22.86} & \best{22.08} & \best{20.95} & \best{20.54} & \best{20.21} & \best{18.77} & \best{17.70} & \best{4.6}                            \\ \midrule
                             & MetaSR~\cite{hu2019metasr}     &29.33 &27.03 &25.66 &24.69 &23.94 &22.82 &22.39 &22.01 &20.42 &19.25    &\second{123.5}                           \\
                             & LIIF~\cite{chen2021liif}       & 29.27 & 26.99 & 25.60  & 24.63 & 23.89 & 22.77 & 22.34 & 21.94 & 20.36 & 19.19 &    480.6                           \\
                             
                             & LTE~\cite{lee2022lte}        & 29.50  & 27.20  & 25.81 & 24.84 & 24.09 & 22.94 & 22.50  & 22.12 & 20.50  & \second{19.31} &1407.5                              \\
                             & SRNO~\cite{wei2023super}        &    29.42	  & 27.12  & 	25.74	  & 24.77	  & 24.03	  & 22.90  & 	22.46	  & 22.06	  & 20.47	  & 19.27    & 390.9                                                         \\
                             & CiaoSR~\cite{cao2023ciaosr}  &   \second{29.59}  &\second{27.28} &	\second{25.89}	 &\second{24.91}	 &\second{24.15}	 & \second{22.99}	 &\second{22.54}	 &\second{22.16}	 &\second{20.50}	 &19.30    & 1857.8                           \\
                             & MambaSR~\cite{yan2024mambasr} &29.36 &27.08 &25.70 &24.74 &23.99 &22.87 &22.44 &22.05 &20.46 &19.27 &398.3
                             \\
                             & GaussianSR~\cite{GaussianSR} & 29.03 & 26.73 & 25.29 & 24.23 & 23.44 & 22.26 & 21.81 & 21.42 & 19.90  & 18.76 & 4962.8                              \\
\multirow{-8}{*}{DIV2K~\cite{Agustsson_2017_CVPR_Workshops}}      & Ours       & \best{29.80}  & \best{27.47} & \best{26.07} & \best{25.08} & \best{24.33} & \best{23.18} & \best{22.74} & \best{22.35} & \best{20.68} & \best{19.45}    & \best{4.7}                           \\ \midrule
                             & MetaSR~\cite{hu2019metasr}    &26.54 &24.64 &23.54 &22.79 &22.24 &21.42 &21.09 &20.80 &19.62 &18.68 & \second{50.4}                               \\
                             & LIIF~\cite{chen2021liif}      & 26.49 & 24.59 & 23.49 & 22.75 & 22.21 & 21.40  & 21.09 & 20.75 & 19.59 & 18.65  &226.4                             \\
                             
                             & LTE~\cite{lee2022lte}      & 26.73 & 24.78 & 23.65 & 22.88 & 22.33 & \second{21.48} & \second{21.15} & \second{20.85} & \second{19.66} & \second{18.71}        & 451.5                       \\
                             & SRNO~\cite{wei2023super}    &  26.65	 &24.72	 &23.61 &	22.85 &	22.30 &	21.45	 & 21.12  &	20.83	       &  19.64     &  18.69       & 163.6                                                      \\
                             & CiaoSR~\cite{cao2023ciaosr}  & \second{26.80} & \second{24.84} & \second{23.69} & \second{22.92} & \second{22.35} & {21.48} & 21.14 & 20.84 & 19.63 & 18.67    &1289.3                           \\
                             & MambaSR~\cite{yan2024mambasr} &26.62 &24.69 &23.59 &22.83 &22.28 &21.44 &21.11 &20.82 &19.64 &18.70 &197.7
                             \\
                             & GaussianSR~\cite{GaussianSR}& 26.25 & 24.39 & 23.28 & 22.49 & 21.92 & 21.06 & 20.74 & 20.45 & 19.29 & 18.38 & 1284.3                              \\
\multirow{-8}{*}{LSDIR~\cite{li2023lsdir}}      & Ours      & \best{27.14} & \best{25.07} & \best{23.91} & \best{23.13} & \best{22.54} & \best{21.89} & \best{21.35} & \best{21.06} & \best{19.79} & \best{18.82}   & \best{4.6}                            \\ \bottomrule
\end{tabular}
}
\vspace{-4mm}
\label{tab:PSNR_comp}
\end{table*}

\subsection{\NAMECov}
Directly learning Gaussian covariance parameters remains challenging due to their unknown range and sensitive space. To address this issue, we propose a novel \NAMECov, which leverages the deep Gaussian prior (DGP) to construct a set of pre-defined Gaussian kernels. This approach simplifies the complex task of directly learning covariance parameters into learning a set of weighting coefficients to combine the pre-defined kernels and represent the target kernel, as shown in Figure~\ref{fig:fig3}.

Specifically, using the DGP, we sample the three covariance parameters \(\sigma_x^2\), \(\sigma_y^2\), and \(\rho\) from the corresponding distributions of the DGP. These parameters are then used to construct a dictionary of \(N\) pre-defined Gaussian kernels. The sampling process is expressed as:
\begin{equation}
\sigma_{i,x}^2, \sigma_{i,y}^2 \sim \mathcal{P}(\sigma_x^2), \mathcal{P}(\sigma_y^2);\rho_i\sigma_{i,x}\sigma_{i,y} \sim \mathcal{P}(\rho\sigma_{x}\sigma_{y}),
\end{equation}
to construct pre-defined Gaussian kernels $\mathcal{K}$:
\begin{equation}
\mathcal{K} = \{ G_i( \begin{bmatrix}
\sigma_{i,x}^2 & \rho_i\sigma_{i,x}\sigma_{i,y} \\
\rho_i\sigma_{i,x}\sigma_{i,y} & \sigma_{i,y}^2
\end{bmatrix} ) \}_{i=1}^N.
\end{equation}
These candidate covariance kernels cover the majority of types and ranges commonly observed in natural images, providing valuable prior information to facilitate network convergence.
We then extract features \(\mathcal{F}_{\text{LR}}\) from the input LR image \(\mathbf{I}_{\text{LR}}\) using the backbone encoder \(\mathbb{E}\):
\begin{equation}
\mathcal{F}_{\text{LR}} = \mathbb{E}(\mathbf{I}_{\text{LR}}).
\end{equation}
To adaptively generate the target covariance kernel, we introduce an adaptive weighting mechanism that learns a set of weights \(\mathbf{W} = \{ w_i \}_{i=1}^N\) based on the extracted features. These weights are computed by the adaptive weighting module \(\mathcal{M}_{\text{weight}}\), which operates as follows:
\begin{equation}
\mathbf{W} = \text{Softmax}(\mathcal{M}_{\text{weight}}(\mathcal{F}_{\text{LR}})).
\end{equation}
The adaptive weighting module \(\mathcal{M}_{\text{weight}}\) is implemented using several layers of convolutional neural networks (CNNs). Finally, each target kernel is generated by performing a weighted combination of the pre-defined kernels in the dictionary:
\begin{equation}
G_{\text{target}} = \sum_{i=1}^N w_i \cdot {G}_i.
\end{equation}
Through the proposed method, we achieve effective optimization of Gaussian covariance, providing stronger prior knowledge and avoiding the local optima observed in method (a), as demonstrated in the ablation study in Section~\ref{sec:fail_solution}.

\subsection{\NAMEPos}
The position parameters are also critical for Gaussian kernels, as they determine their locations in the 2D space. Directly learning these positions is highly challenging, as demonstrated in Section~\ref{sec:fail_solution}. Since each LR pixel typically corresponds to multiple pixels in the HR image, a straightforward solution is to fix the positions at the centers of the LR pixels. While this strategy simplifies the optimization process, it significantly limits the model's representational capacity, making it difficult to adaptively learn the position distribution based on image content.

To address the above issues, we propose a novel method, \NAMEPos (APD), which not only ensures efficient optimization but also improves representational capacity by allowing the model to adaptively learn positions, as shown in Figure~\ref{fig:fig3}. Specifically, we use the center positions of LR pixels as the initialized positions ${P}_\text{init}$ and further introduce a dynamic offset mechanism, which learns a dynamic offset from LR features $\mathcal{F}_\text{LR}$ by $\mathcal{M}_{\text{pos}}$ model to adjust the spatial positions. Here, we set the offset range from \(-1 \sim 1\) by the Tanh activate function and add the offset $P_\text{{off}}$ to the initialized LR center positions to obtain the final positions $P_\text{{final}}$, as expressed by the following equation:
\begin{align}
P_{\text{off}} &= \text{Tanh}(\mathcal{M}_{\text{pos}}(\mathcal{F}_{\text{LR}})), \\
P_{\text{final}} &= P_{\text{init}} + P_{\text{off}}.
\end{align}
where $\mathcal{M}_\text{pos}$ is implemented using five multilayer perceptron layers. This $P_\text{off}$ enables the network to adaptively learn kernel positions based on image content, resulting in denser kernel placement in regions with richer textures and enhancing the network's performance, as demonstrated in Figure~\ref{fig:fig3}, Section~\ref{sec:ab_study}, Appendix Section~\ref{sec:offset}.

In addition, since the RGB is range from $0$ to $1$ and is relatively easy to optimize, we introduce a simple Color Gaussian Mapping (CGM) to learn the RGB parameters. Specifically, this mapping is implemented using \(5\) multilayer perceptron (MLP) layers applied to \(\mathcal{F}_{\text{LR}}\). In summary, the above three components construct our proposed {ContinuousSR} framework, as shown in Figure~\ref{fig:fig3}.

\begin{table*}[t]
\centering
\caption{Performance comparison of the Urban100 benchmark on SSIM, FID, and DISTS metrics.}
\vspace{-3mm}
\scalebox{0.90}{
\begin{tabular}{l|l|cccccccccc}
\toprule
Metrics & Methods    & $\times$4     & $\times$6     & $\times$8     & $\times$10    & $\times$12    & $\times$16    & $\times$18    & $\times$20    & $\times$32    & $\times$48    \\ \midrule
                                                      & LIIF~\cite{chen2021liif}       &0.7911 &0.6861 &0.6148 &0.5642 &0.5270 &0.4790 &0.4617 &0.4503 &0.4106 
                                                      &0.3918\\
                                                      & LTE~\cite{lee2022lte}        & 0.8069 & 0.7045 & 0.6321 & 0.5810 & 0.5422 & 0.4900 & 0.4710 & 0.4588 & 0.4145 & 0.3931 \\
                                                      & GaussianSR ~\cite{GaussianSR} & 0.7751 & 0.6633 & 0.5867 & 0.5334 & 0.4967 & 0.4521 & 0.4369 & 0.4277 & 0.3969 & 0.3835 \\
                                                      & CiaoSR   ~\cite{cao2023ciaosr}  & \second{0.8110} & \second{0.7126} & \second{0.6415} & \second{0.5887} & \second{0.5503} & \second{0.4974} & \second{0.4777} & \second{0.4637} & \second{0.4168} & \second{0.3921} \\
\multirow{-5}{*}{SSIM$\uparrow$}                                & Ours       & \best{0.8292} & \best{0.7343} & \best{0.6624} & \best{0.6089} & \best{0.5683} & \best{0.5097} & \best{0.4893} & \best{0.4746} & \best{0.4216} & \best{0.3958} \\ \midrule
                                                      & LIIF    ~\cite{chen2021liif}   &0.1611 &0.2178 &0.2589 &0.2926 &0.3209 &0.3659 &0.3835 &0.3990 &0.4678 &0.5322       \\
                                                      & LTE ~\cite{lee2022lte}       & 0.1570 & 0.2126 & 0.2541 & 0.2872 & 0.3157 & 0.3611 & 0.3799 & 0.3960 & 0.4695 & 0.5362 \\
                                                      & GaussianSR~\cite{GaussianSR}  & 0.1740 & 0.2374 & 0.2890 & 0.3296 & 0.3631 & 0.4109 & 0.4302 & 0.4466 & 0.5157 & 0.5713 \\
                                                      & CiaoSR  ~\cite{cao2023ciaosr}   & \second{0.1533} & \second{0.2074} & \second{0.2453} & \second{0.2771} & \second{0.3049} & \second{0.3510} & \second{0.3701} & \second{0.3863} & \second{0.4513} & \best{0.4998} \\
\multirow{-5}{*}{DISTS$\downarrow$}                               & Ours       & \best{0.1356} & \best{0.1901} & \best{0.2299} & \best{0.2601} & \best{0.2860} & \best{0.3324} & \best{0.3504} & \best{0.3670} & \best{0.4439} & \second{0.5144} \\ \midrule
                                                       & LIIF    ~\cite{chen2021liif}   &4.76 &24.87 &50.05 &77.54 &102.47 &145.25 &164.41 &179.44 &256.95 &311.09       \\
                                                      & LTE  ~\cite{lee2022lte}      & 3.84   & 21.01  & 45.15  & 70.56  & 92.85  & 136.54 & 156.24 & 170.92 & 253.52 & 296.27 \\
                                                      & GaussianSR~\cite{GaussianSR} & 5.64   & 29.30  & 57.00  & 89.25  & 120.02 & 166.20 & 181.59 & 202.18 & 264.27 & 315.97 \\
                                                      & CiaoSR ~\cite{cao2023ciaosr}    & \second{3.74}   & \second{20.48}  & \second{43.25}  & \second{58.60}  & \second{92.58}  & \second{133.77} & \second{151.70} & \second{168.89} & \second{247.84} & \second{294.49} \\
\multirow{-5}{*}{FID$\downarrow$}                                 & Ours       & \best{2.91}   & \best{16.50}  & \best{37.09}  & \best{58.83}  & \best{78.72}  & \best{116.30} & \best{130.32} & \best{143.52} & \best{216.21} & \best{281.27} \\ \bottomrule
\end{tabular}
}
\label{tab:SSIM_FID_comp}
\end{table*}
\begin{figure*}[t]
  \centering
  \includegraphics[width=0.98\linewidth]{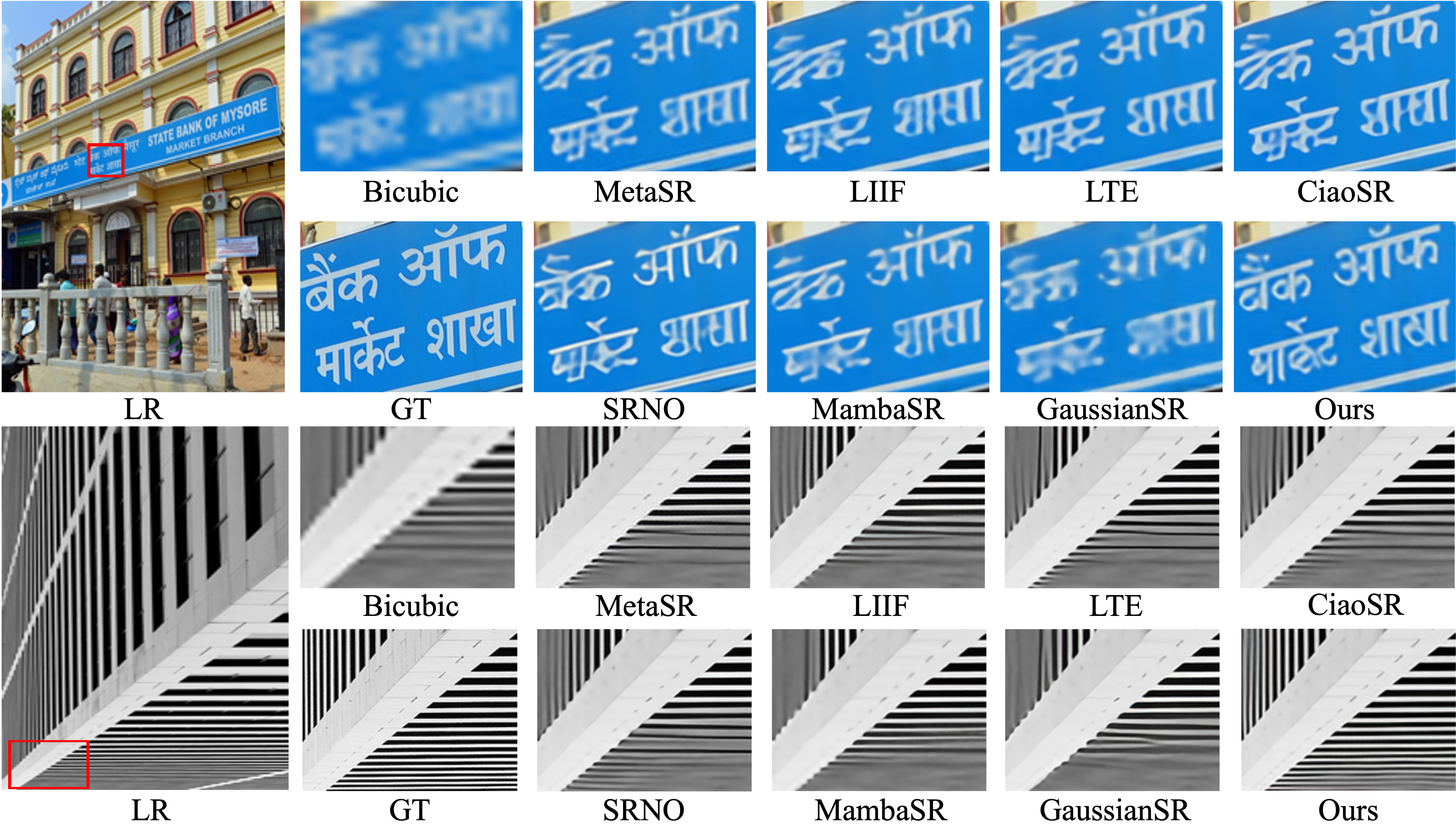}
  \vspace{-2mm}
\caption{Qualitative comparison. The visual quality of our method outperforms existing methods. Please zoom in for a better view.}
\vspace{-2mm}
  \label{fig:vis1}
  
\end{figure*}

\section{Experiment and Analysis}
\label{sec:exp}
\subsection{Experiment Setting}
{\bf Datasets.} We use the commonly employed DF2K high-quality dataset~\cite{wang2021real} as HR images, which are degraded using bicubic to generate LR for training. For evaluation, we adopt Set5~\cite{bevilacqua2012low}, Set14~\cite{zeyde2012single}, B100~\cite{martin2001database}, Urban100~\cite{huang2015single},
Manga109~\cite{matsui2017sketch}, DIV2K validation~\cite{Agustsson_2017_CVPR_Workshops} and LSDIR~\cite{li2023lsdir}. \\
{\bf Evaluation metrics.} Following previous work~\cite{chen2021liif,lee2022lte}, we use PSNR, SSIM~\cite{SSIM}, FID~\cite{FID}, and DISTS~\cite{DISTS} for evaluation. Note that the PSNR/SSIM value is calculated on the RGB channels for the DIV2K validation set and on the Y channel (\ie, luminance) of the transformed YCbCr space for the other benchmark test sets.\\
{\bf Implementation details.} Following previous works~\cite{chen2021liif,lee2022lte}, we adopt the same way to generate paired images for training. Specifically, initially, we crop image patches of size 256 $\times$ 256 as ground truth. Then, we use bicubic downsampling to generate corresponding LR images, and the downsampling scaling factor is sampled from a uniform distribution U(4, 8). We employ SwinIR~\cite{liang2021swinir} and HAT~\cite{chen2023hat} as backbones. Adam~\cite{Adam} is used as the optimizer, with the initial learning rate setting to 1e-4 and decaying by a factor of 0.5 every 100 epochs. We utilize the L1 loss~\cite{chen2021liif} and frequency loss~\cite{focal} for training, with a total batch size of 64 and 1000 training epochs on 8 V100 GPUs.\\
{\bf Compared methods.} We compare with nine state-of-the-art and popular models: MetaSR~\cite{hu2019metasr}, LIIF~\cite{chen2021liif}, LTE~\cite{lee2022lte}, ITSRN~\cite{yang202itsrn}, SRNO~\cite{wei2023super}, CiaoSR~\cite{cao2023ciaosr}, MambaSR~\cite{yan2024mambasr}, GaussianSR~\cite{GaussianSR} and GSASR~\cite{chen2025generalized}. The best version from thier official code is used for comparison. Details are provided in {Appendix} Section~\ref{sec:detail_of_compared_methods}.
\begin{table}[t]
\centering
\caption{Memory usage (G) comparison with different methods.}
\vspace{-1mm}
\scalebox{0.9}{
\begin{tabular}{l|ccccc}
\toprule
Memory Usage & $\times$4 & $\times$6 & $\times$8 & $\times$12 & $\times$16 \\
\midrule
LIIF   & 4.12  & 6.49  & 9.79  & 19.27 & \textbf{OOM} \\
CiaoSR & 12.17 & 22.83 & \textbf{OOM}   & \textbf{OOM}   & \textbf{OOM} \\
Ours   & \best{2.48}  & \best{2.49}  & \best{2.50}  & \best{2.52}  & \best{2.54} \\
\bottomrule
\end{tabular}
}
\vspace{-1mm}
\label{tab:memory_usage}
\end{table}
\subsection{Quantitative and Qualitative Results}
{\bf Quantitative comparisons.} As shown in Table~\ref{tab:PSNR_comp} and Table~\ref{tab:SSIM_FID_comp}, our method achieves the best performance compared to existing approaches across all evaluation metrics and benchmarks. For example, on the Urban100 dataset, our method surpasses the current state-of-the-art (CiaoSR) by 0.80 dB in the \(\times 4\), representing a substantial improvement. Similarly, in terms of SSIM and FID, our method achieves further gains in the \(\times 4\) scenario, surpassing the current state-of-the-art by 0.0172 and 0.83, respectively. These results demonstrate the effectiveness and superiority of the proposed method.\\
{\bf Complexity comparisons.} We present comparisons of runtime and memory usage. Specifically, we evaluate the average runtime across 45 different scales, ranging from $\times$4 to $\times$48. As shown in Table~\ref{tab:PSNR_comp}, our method significantly outperforms existing methods in terms of speed. For instance, it surpasses the current state-of-the-art method (CiaoSR) by nearly 280 times on the LSDIR dataset. Moreover, we also provide comparisons of memory usage. Specifically, we set the input size to 48$\times$48, disable the tiling strategy, and test the memory usage under different scales. As shown in Table~\ref{tab:memory_usage}, thanks to our efficient pipeline design, our method maintains minimal computational overhead across different scales. In contrast, existing methods, such as LIIF and CiaoSR, fail to handle larger scales and encounter OOM (out of memory) on V100 GPUs. \\
{\bf Qualitative comparisons.} We present qualitative comparisons, as shown in Figure~\ref{fig:vis1}. Compared to existing methods, our approach reconstructs sharper and more visually pleasing details that are consistent with the ground truth (GT). For instance, in the bottom part of Figure~\ref{fig:vis1}, our method effectively reconstructs the texture details inside the building. This highlights the superiority of our method in generating realistic and perceptually satisfying results. \\
More visual comparisons, user studies, benchmark results, FLOPs comparisons, and details are provided in Appendix Sections~\ref{sec:Addtional_Comparison}, ~\ref{sec:user_study}, and~\ref{sec:addtional_visual_comp}.

\subsection{Ablation Study}
\label{sec:ab_study}
In this section, we present ablation studies on our proposed APD, DDCW, and DGP using the Urban100 $\times4$ dataset. Specifically, we independently remove APD and DDCW, as illustrated in the right part of Table~\ref{tab:ablation_studies}. The exclusion of these modules significantly exacerbates the optimization difficulty, resulting in a considerable decline in PSNR. Then, we validate the effectiveness of $P_\text{init}$ and $P_\text{off}$ in APD. As shown in the middle of Table~\ref{tab:ablation_studies}, the results demonstrate that the model achieves the best representational capacity and performance when both components are employed. Finally, we evaluate the effectiveness of DGP in DDCW $\mathcal{K}_{DCP}$. Specifically, we remove the DGP and separately modify the covariance range to [0,1] and [0,10], using uniform sampling to construct $\mathcal{K}_{1}$ and $\mathcal{K}_{2}$. As shown in the right in Table~\ref{tab:ablation_studies}, incorporating DGP provides a better basis function, thereby enhancing performance. More ablation studies are provided in {Appendix Sections \ref{sec:addtional_ab_study}}.

\begin{table}[t]
    \centering
    \setlength{\tabcolsep}{3pt} 
    \caption{Ablation studies on proposed APD, DDCW, and DGP.}
    \vspace{-1mm}

\scalebox{0.9}{   \begin{tabular}{c|c|c}
        \toprule

        \begin{tabular}{ccc}
            {DDCW} & {APD} & PSNR \\
            \midrule
            \checkmark &        & 10.5\\
            & \checkmark        & 12.3 \\
            \checkmark & \checkmark & \best{28.2} \\
        \end{tabular}
        &
        \begin{tabular}{ccc}
            $P_\text{init}$ & $P_\text{off}$ & PSNR \\
            \midrule
            \checkmark &        & 27.8 \\
            & \checkmark        & 10.5 \\
            \checkmark & \checkmark & \best{28.2} \\
        \end{tabular}
        &
        \begin{tabular}{cc}
            $K$  & PSNR \\
            \midrule
            $\mathcal{K}_{1}$     & 27.7 \\
            $\mathcal{K}_{2}$       & 27.1 \\
            $\mathcal{K}_{DCP}$ & \best{28.2} \\
        \end{tabular} \\
        \bottomrule
    \end{tabular}
    \label{tab:ablation_studies}
    }
    \vspace{-1mm}

\end{table}
\begin{table}[t]
\centering
\caption{Performance comparison under more challenging low-resolution and rainy conditions.}
\scalebox{0.90}{
\begin{tabular}{l|cccccc}
\toprule
Methods              & $\times$4    & $\times$5    & $\times$6    & $\times$7    & $\times$8    \\ \midrule
LIIF                 & 24.04 & 23.53 & 23.09 & 22.70  & 22.39 \\
GaussianSR           & 24.04 & 23.51 & 23.08 & 22.69 & 22.38 \\
CiaoSR               & 23.84 & 23.45 & 23.01 & 22.66 & 22.33 \\
Ours                 & \best{24.51} & \best{23.95} & \best{23.48} & \best{23.07} & \best{22.76} \\ \bottomrule
\end{tabular}
}
\label{tab:ASSR_Derain}
\end{table}

\section{Future Work}
It is well known that, in real-world scenarios, image degradation is not limited to low resolution but often includes other types of degradation, such as rain and noise. The goal of low-level vision is to remove these degradations while enhancing image resolution and quality. To this end, we evaluate our method on Rain200H~\cite{yang2017deep}, simulating low-resolution rainy images with bicubic downsampling. We compare our approach with three existing state-of-the-art methods to validate its effectiveness. As shown in Table~\ref{tab:ASSR_Derain}, our method removes rain degradations more effectively while enhancing resolution and details, outperforming existing methods. This demonstrates the potential of our approach for other low-level vision tasks. In future work, we aim to extend it to more tasks to further enhance its applicability.

To provide a comprehensive understanding of our methods, we include detailed explanations, additional comparisons, analyses, limitations, future work, and extensive visual examples in the \textbf{Appendix}, showcasing superiority.

\section{Conclusion}
We introduce \NAME, a novel Pixel-to-Gaussian paradigm designed for fast and high-quality arbitrary-scale super-resolution. By explicitly reconstructing 2D continuous HR signals from LR images using Gaussian Splatting, \NAME significantly improves both efficiency and performance. Through statistical analysis, we uncover the Deep Gaussian Prior (DGP) and propose a DGP-driven Covariance Weighting mechanism along with an Adaptive Position Drifting strategy. These innovations improve the quality and fidelity of the reconstructed Gaussian fields. Experiments on seven popular benchmarks demonstrate that our method outperforms state-of-the-art methods in both quality and speed, achieving a {19.5×} speed improvement and {0.90dB} PSNR improvement, making it a promising solution for ASSR tasks.

{
    \small
    \bibliographystyle{ieeenat_fullname}
    \bibliography{main}
}

\clearpage
\setcounter{page}{1}
\maketitlesupplementary

\begin{table*}[h!]
\centering
\caption{Performance comparison with existing methods using the same SwinIR~\cite{liang2021swinir} backbone on the Set5~\cite{bevilacqua2012low}, Set14~\cite{zeyde2012single}, B100~\cite{martin2001database}, Urban100~\cite{huang2015single} and
Manga109~\cite{matsui2017sketch} datasets. Table performance is referred to in \cite{cao2023ciaosr}.}
\setlength{\tabcolsep}{3pt} 
\scalebox{0.90}{\begin{tabular}{l|cccccccc}
\toprule
Dataset & Scale      & SwinIR~\cite{liang2021swinir}   & MetaSR~\cite{hu2019metasr}   & LIIF~\cite{chen2021liif}     & ITSRN~\cite{yang202itsrn}    & LTE~\cite{lee2022lte}      & CiaoSR~\cite{cao2023ciaosr}   & Ours     \\ \midrule
\multirow{4}{*}{Set5~\cite{bevilacqua2012low}}   
        & $\times$4  & 32.72    & 32.47    & 32.73    & 32.63    & 32.81    & \second{32.84}    & \best{32.93}    \\
        & $\times$6  & -        & 29.09    & 29.46    & 29.31    & 29.50    & \second{29.62}    & \best{29.67}    \\
        & $\times$8  & -        & 27.02    & 27.36    & 27.24    & 27.35    & \second{27.45}    & \best{27.55}    \\
        & $\times$12 & -        & 24.82    & -        & 24.79    & -        & \second{24.96}    & \best{25.18}    \\ \midrule
\multirow{4}{*}{Set14~\cite{zeyde2012single}}  
        & $\times$4  & 28.94    & 28.85    & 28.98    & 28.97    & 29.06    & \second{29.08}    & \best{29.18}    \\
        & $\times$6  & -        & 26.58    & 26.82    & 26.71    & 26.86    & \second{26.88}    & \best{26.96}    \\
        & $\times$8  & -        & 25.09    & 25.34    & 25.32    & 25.42    & \second{25.42}    & \best{25.54}    \\
        & $\times$12 & -        & 23.33    & -        & 23.30    & -        & \second{23.38}    & \best{23.55}    \\ \midrule
\multirow{4}{*}{B100~\cite{martin2001database}} 
        & $\times$4  & 27.83    & 27.75    & 27.84    & 27.85    & 27.86    & \second{27.90}    & \best{27.94}    \\
        & $\times$6  & -        & 25.94    & 26.07    & 26.05    & 26.09    & \second{26.13}    & \best{26.18}    \\
        & $\times$8  & -        & 24.87    & 25.01    & 24.96    & 25.03    & \second{25.07}    & \best{25.13}    \\
        & $\times$12 & -        & 23.59    & -        & 23.57    & -        & \second{23.68}    & \best{23.76}    \\ \midrule
\multirow{4}{*}{Urban100~\cite{huang2015single}} 
        & $\times$4  & 27.07    & 26.76    & 27.15    & 27.12    & 27.24    & \second{27.42}    & \best{27.51}    \\
        & $\times$6  & -        & 24.16    & 24.59    & 24.50    & 24.62    & \second{24.84}    & \best{24.93}    \\
        & $\times$8  & -        & 22.75    & 23.14    & 23.06    & 23.17    & \second{23.34}    & \best{23.45}    \\
        & $\times$12 & -        & 21.31    & -        & 21.34    & -        & \second{21.60}    & \best{21.72}    \\ \midrule
\multirow{4}{*}{Manga109~\cite{matsui2017sketch}} 
        & $\times$4  & 31.67    & 31.37    & 31.71    & 31.74    & 31.79    & \second{31.91}    & \best{32.19}    \\
        & $\times$6  & -        & 27.29    & 27.69    & 27.72    & 27.83    & \second{28.01}    & \best{28.25}    \\
        & $\times$8  & -        & 24.96    & 25.28    & 25.23    & 25.42    & \second{25.61}    & \best{25.80}    \\
        & $\times$12 & -        & 22.35    & -        & 22.47    & -        & \second{22.79}    & \best{22.96}    \\ 
\bottomrule
\end{tabular}
}
\label{tab:SwinIR}
\end{table*}

\begin{table*}[h!]
\centering
    \setlength{\tabcolsep}{3pt} 

\caption{Comparison of PSNR (dB), FLOPs (G), and running time (ms) on the Manga109 dataset.}
\begin{tabular}{lccccccccc}
\toprule
 & \multicolumn{3}{c}{$\times$4} & \multicolumn{3}{c}{$\times$6} & \multicolumn{3}{c}{$\times$8} \\
\cmidrule(lr){2-4} \cmidrule(lr){5-7} \cmidrule(lr){8-10}
 & PSNR & FLOPs & Times & PSNR & FLOPs & Times & PSNR & FLOPs & Times \\
\midrule
SwinIR-LIIF           & 31.71 & 365.11 & 150 & 27.69 & 289.77 & 141 & 25.28 & 271.35 & 137 \\
SwinIR-CiaoSR         & 31.91 & 1949.71 & 319 & 28.01 & 1275.95 & 256 & 25.61 & 1048.08 & 235 \\
SwinIR-ContinuousSR   & \best{32.19} & \best{280.99} & \best{136} & \best{28.25} & \best{125.63} & \best{112} & \best{25.80} & \best{79.21} & \best{99} \\
\bottomrule
\end{tabular}
\label{tab:FLOPs}
\end{table*}

\begin{table*}[h]
\centering
\caption{Comparison of PSNR on the COZ dataset.}
\label{tab:coz_psnr}
\begin{tabular}{l|ccccccccc}
\toprule
\textbf{COZ} & MetaSR & LIIF & LTE & LINF & SRNO & LIT & CiaoSR & LMI & Ours \\
\midrule
PSNR         & 24.39  & 24.39 & 24.4 & 24.32 & 24.4  & 24.36 & 24.38  & 24.48 & \textbf{24.68} \\
\bottomrule
\end{tabular}
\label{tab:COZ}
\end{table*}

\begin{table*}[t!]
\centering
    \setlength{\tabcolsep}{3pt} 

\caption{LPIPS$\downarrow$ comparison for Urban100 dataset across different methods.}
\begin{tabular}{l|cccccccc}
\toprule
LPIPS$\downarrow$       & MetaSR & LIIF   & MambaSR & LTE    & SRN0   & CiaoSR & GaussianSR & Ours   \\ \midrule
Urban100    & 0.1989 & 0.2080  & 0.2073  & 0.1934 & 0.1991 & 0.1872 & 0.2285     & \best{0.1803} \\ 
\bottomrule
\end{tabular}
\label{tab:LPIPS_Urban100}
\end{table*}

\section{Details of Compared Methods}
\label{sec:detail_of_compared_methods}

To validate the effectiveness of our proposed model, we compare it against seven state-of-the-art (SOTA) and widely adopted models: MetaSR~\cite{hu2019metasr}, LIIF~\cite{chen2021liif}, LTE~\cite{lee2022lte}, SRNO~\cite{wei2023super}, CiaoSR~\cite{cao2023ciaosr}, MambaSR~\cite{yan2024mambasr} and GaussianSR~\cite{GaussianSR}. For a fair comparison, we select the best-performing networks for each method based on their official GitHub repositories. Specifically, we use the MetaSR model based on SwinIR, the LIIF model based on RDN, the LTE model based on SwinIR, the SRNO model based on RDN, the CiaoSR model based on SwinIR, the MambaSR model based on RDN, and the GaussianSR model on EDSR-baseline.

\section{User Study}
\label{sec:user_study}
\begin{figure}[t]
  \centering
  \includegraphics[width=1\linewidth]{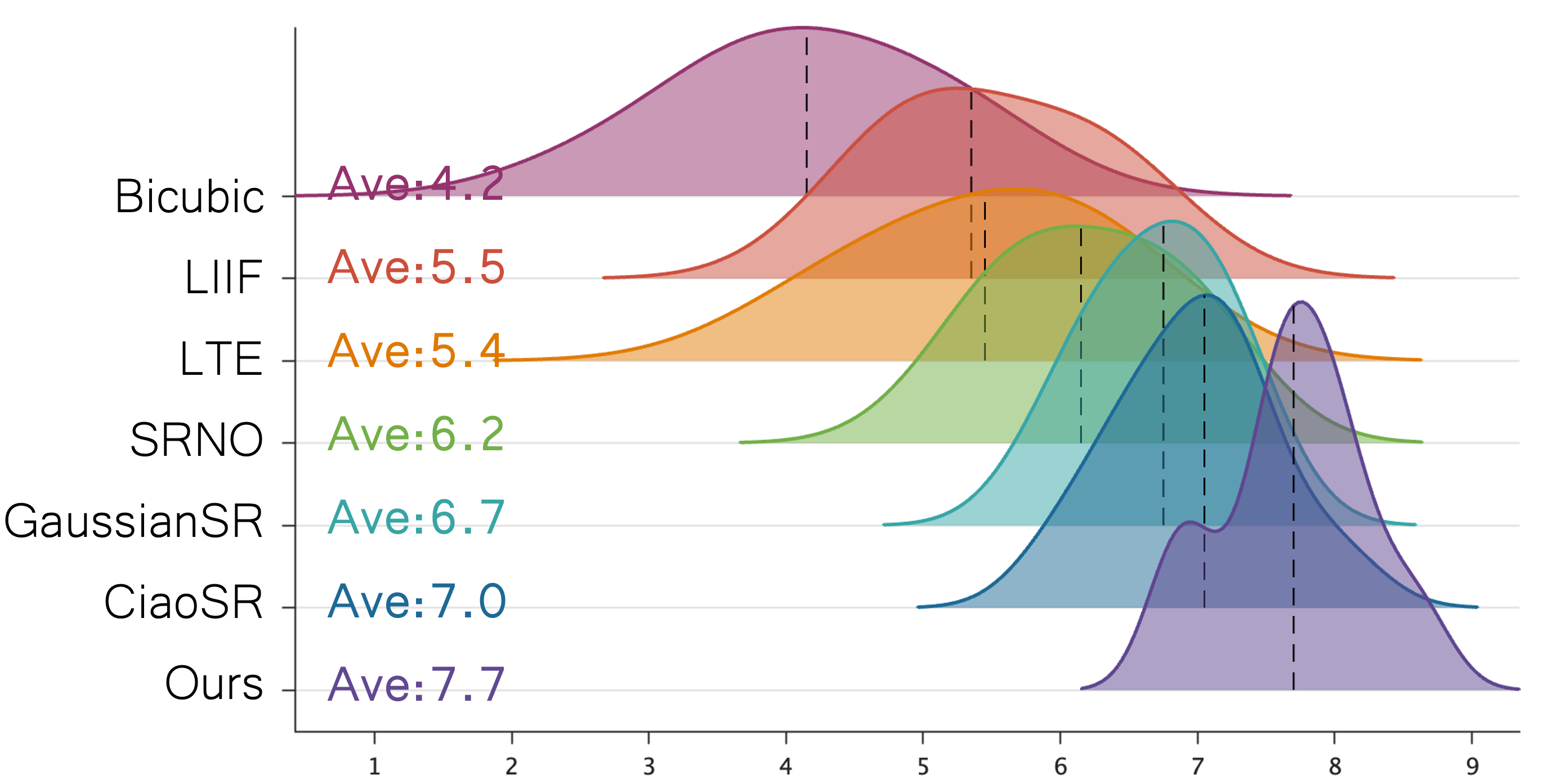}
\caption{User study.}
  \label{fig:user_study}
\end{figure}

To further assess visual quality, we conduct a user study. Ten images are randomly selected from the test datasets, ensuring diversity in image content and complexity. Fifteen participants rate the visual quality of each processed image on a scale from 0 (poor) to 10 (excellent). Each participant evaluates the images independently to ensure unbiased results. As shown in Figure~\ref{fig:user_study}, the results demonstrate that existing methods frequently fail to restore high-quality images, particularly in challenging regions with fine details and textures. This leads to lower user satisfaction, with average scores ranging between 4.2 and 7.0 for most competing methods. In contrast, our method achieves the highest average score of 7.7, significantly outperforming all other approaches. The superior performance of our method demonstrates its ability to produce sharper details, better texture preservation, and visually consistent results. Participants consistently note that our method outperforms others, particularly in challenging regions, further validating its effectiveness and generalization in restoring high-quality images.

\section{Addtional Comparison}
\label{sec:Addtional_Comparison}
{\bf More Benchmarks.} To further demonstrate the superiority of our proposed method, we conduct experiments to compare its performance against existing methods using the same SwinIR~\cite{liang2021swinir} backbone on the Set5~\cite{bevilacqua2012low}, Set14~\cite{zeyde2012single}, B100~\cite{martin2001database}, Urban100~\cite{huang2015single} and
Manga109~\cite{matsui2017sketch} datasets. As shown in Table~\ref{tab:SwinIR}, our method still achieves state-of-the-art performance across all benchmarks and scales. \\
{\bf More Comparisons on Real Datasets.} To further demonstrate the superiority of the proposed method in real-world scenarios, we compare it with existing methods on the real dataset COZ~\cite{fu2024continuous} on $\times5$. The results, as shown in Table~\ref{tab:COZ}, indicate that our method consistently outperforms existing approaches in real-world scenarios, validating the superior generalization ability of our method to real-world data.\\
{\bf More Complexity Comparisons.} Furthermore, we also provide comparisons of FLOPs and inference time at a single scale on the Manga109 dataset. Specifically, considering that the image shapes in the original dataset may cause other methods to run out of memory, we fix the GT shape to 288 and evaluate the FLOPs and inference time at different scales. As shown in Table~\ref{tab:FLOPs}, our method not only achieves the best performance in terms of PSNR but also maintains the lowest FLOPs and inference time at a single scale, significantly outperforming the current SOTA method, CiaoSR. Moreover, Table~1 in the main text further demonstrates our superiority in total runtime across multiple scales. These results fully validate the efficiency and superiority of the proposed method.\\
{\bf More Performance Metrics.} In the main text, we have provided PSNR, SSIM, FID, and DISTS metrics to demonstrate the superiority of the proposed method. Here, we further present a comparison of LPIPS performance on the Urban100 $\times$ 4 dataset. As shown in Table~\ref{tab:LPIPS_Urban100}, our method achieves the best performance in terms of LPIPS. This further validates the superiority of the proposed method in perceptual quality.\\
{\bf More Compared Method.} In the main text, we have compared our proposed method with 9 existing methods to demonstrate its superiority. Additionally, we include a comparison with GaussianImage~\cite{zhang2024gaussianimage}. Specifically, we conduct experiments on the Set5 and Set14 datasets under the $\times$4 scenario. Since GaussianImage is an optimization-based end-to-end algorithm, we allow this method to optimize on LR inputs and adjust the Gaussian mapping scale to perform super-resolution for comparison. As shown in Table~\ref{tab:GaussianImage}, this method fails to learn the mapping from LR to HR, resulting in poor performance. Furthermore, it is worth noting that GaussianImage requires nearly 1 minute of optimization per scene on a V100 GPU, which is impractical for real-world applications. \\
{\bf More Compared Methods with GS.} Several recent GS-based ASSR methods have been proposed, such as GaussianSR~\cite{GaussianSR} and GSASR~\cite{chen2025generalized}. GaussianSR has already been thoroughly analyzed and compared in the main text. Here, we focus on analyzing and comparing GSASR. Although GSASR has made notable progress, it is still constrained by inefficiencies caused by multiple upsampling and decoding processes across different scales. Furthermore, GSASR performs GS in the feature and image space, which makes it struggle to ensure the continuity of reconstructed images across different scales, leading to low performance. In contrast, our method leverages 2D GS modeling to reconstruct continuous HR images, enabling both fast and high-quality ASSR. Although the GSASR method has not been open-sourced, we still compare our method against the performance reported in its paper. For example, on the LSDIR benchmark, our method achieves a performance of 27.14 dB at $\times 4$, significantly surpassing GSASR's best reported performance of 26.73 dB. This demonstrates the superiority of our method in terms of performance. Moreover, in terms of speed, our method requires only 1 ms to generate high-quality HR images across different scales, whereas GSASR takes approximately 91-1573 ms. This further highlights the ultra-fast speed of our proposed method.\\
{\bf More Details in Section 5.2.} In Table~\ref{tab:PSNR_comp}, the Average Time (AT) is calculated by performing super-resolution on LR images across 45 different scales, ranging from \(\times4\) to \(\times48\), and then averaging the total runtime. For each dataset, we select a representative LR shape and downsample it by a factor of 48 to construct the input size for each dataset, ensuring that existing ASSR methods do not encounter out-of-memory (OOM) issues. Specifically, the LR size is \(21 \times 13\) for Urban100, \(42 \times 28\) for DIV2K, and \(29 \times 19\) for LSDIR. As shown in Table~\ref{tab:PSNR_comp}, our method consistently achieves significant speed advantages over existing methods across different datasets and LR shapes.

\begin{table*}[t]
    \centering
    \begin{minipage}{0.24\textwidth}
        \centering
        \caption{Ablation on Dim.}
        \begin{tabular}{c|c|c}
            \toprule
             & PSNR & SSIM \\ \midrule
            3   & 28.14 & 0.8273 \\ 
            256 & 28.17 & 0.8281 \\ 
            512 & 28.22 & 0.8292 \\ \bottomrule
        \end{tabular}
    \label{tab:ab_dim}
    \end{minipage}%
    \begin{minipage}{0.24\textwidth}
        \centering
        \caption{Number of $K$.}
        \begin{tabular}{c|c|c}
            \toprule
             & PSNR & SSIM \\ \midrule
            100 & 28.12 & 0.8277 \\ 
            500 & 28.19 & 0.8286 \\ 
            730 & 28.22 & 0.8292 \\ \bottomrule
        \end{tabular}
            \label{tab:ab_k}

    \end{minipage}%
    \begin{minipage}{0.24\textwidth}
        \centering
        \caption{Number of $N$.}
        \begin{tabular}{c|c|c}
            \toprule
             & PSNR & SSIM \\ \midrule
            1   & 28.01 & 0.8254 \\ 
            4   & 28.22 & 0.8292 \\ 
            9   & 28.18 & 0.8284 \\ \bottomrule
        \end{tabular}
                    \label{tab:ab_n}

    \end{minipage}%
    \begin{minipage}{0.24\textwidth}
        \centering
        \caption{Ablation on $P_\text{off}$.}
        \begin{tabular}{c|c|c}
            \toprule
             & PSNR & SSIM \\ \midrule
            0.5 & 28.03 & 0.8259 \\ 
            1   & 28.22 & 0.8292 \\ 
            2   & 28.17 & 0.8283 \\ \bottomrule
        \end{tabular}
        \label{tab:ab_off}

    \end{minipage}
    
\end{table*}

\begin{table*}[t]
\centering
\caption{Performance comparison of SR and deraining methods under different scaling factors.}
\scalebox{0.90}{
\begin{tabular}{l|l|ccccccccc}
\toprule
Types & Methods              & $\times$4    & $\times$4.5  & $\times$5    & $\times$5.5  & $\times$6    & $\times$6.5  & $\times$7    & $\times$7.5  & $\times$8    \\ \midrule
                             & LIIF+DRSformer       & 17.92 & 16.76 & 15.84 & 15.03 & 14.47 & 14.02 & 13.71 & 13.58 & 13.50  \\
                             & DRSformer+LIIF       & 20.14 & 19.87 & 19.76 & 19.53 & 19.39 & 19.21 & 19.12 & 18.98 & 18.93  \\
                             & GaussianSR+DRSformer & 17.14 & 16.09 & 15.25 & 14.53 & 14.07 & 13.75 & 13.57 & 13.53 & 13.52  \\
                             & DRSformer+GaussianSR & 20.12 & 19.86 & 19.75 & 19.52 & 19.39 & 19.21 & 19.13 & 18.98 & 18.94  \\
                             & CiaoSR+DRSformer     & 19.45 & 18.25 & 17.19 & 16.29 & 15.55 & 14.85 & 14.22 & 13.94 & 13.73  \\
\multirow{-6}{*}{ASSR+Derain}  & DRSformer+CiaoSR     & 19.96 & 19.73 & 19.65 & 19.48 & 19.27 & 19.15 & 19.07 & 18.89 & 18.87  \\ \midrule
                             & LIIF                 & 24.04 & 23.79 & 23.53 & 23.29 & 23.09 & 22.90  & 22.70  & 22.52 & 22.39  \\
                             & GaussianSR           & 24.04 & 23.76 & 23.51 & 23.28 & 23.08 & 22.89 & 22.69 & 22.52 & 22.38  \\
                             & CiaoSR               & 23.84 & 23.66 & 23.45 & 22.23 & 23.01 & 22.83 & 22.66 & 22.50  & 22.33  \\
\multirow{-4}{*}{All in one}    & Ours               & 24.51 & 24.22 & 23.95 & 23.69 & 23.48 & 23.28  & 23.07   & 22.90  & 22.76  \\ \bottomrule
\end{tabular}
}
\label{tab:ASSR_Derain2}
\end{table*}
\begin{figure*}[t]
  \centering
  \includegraphics[width=1\linewidth]{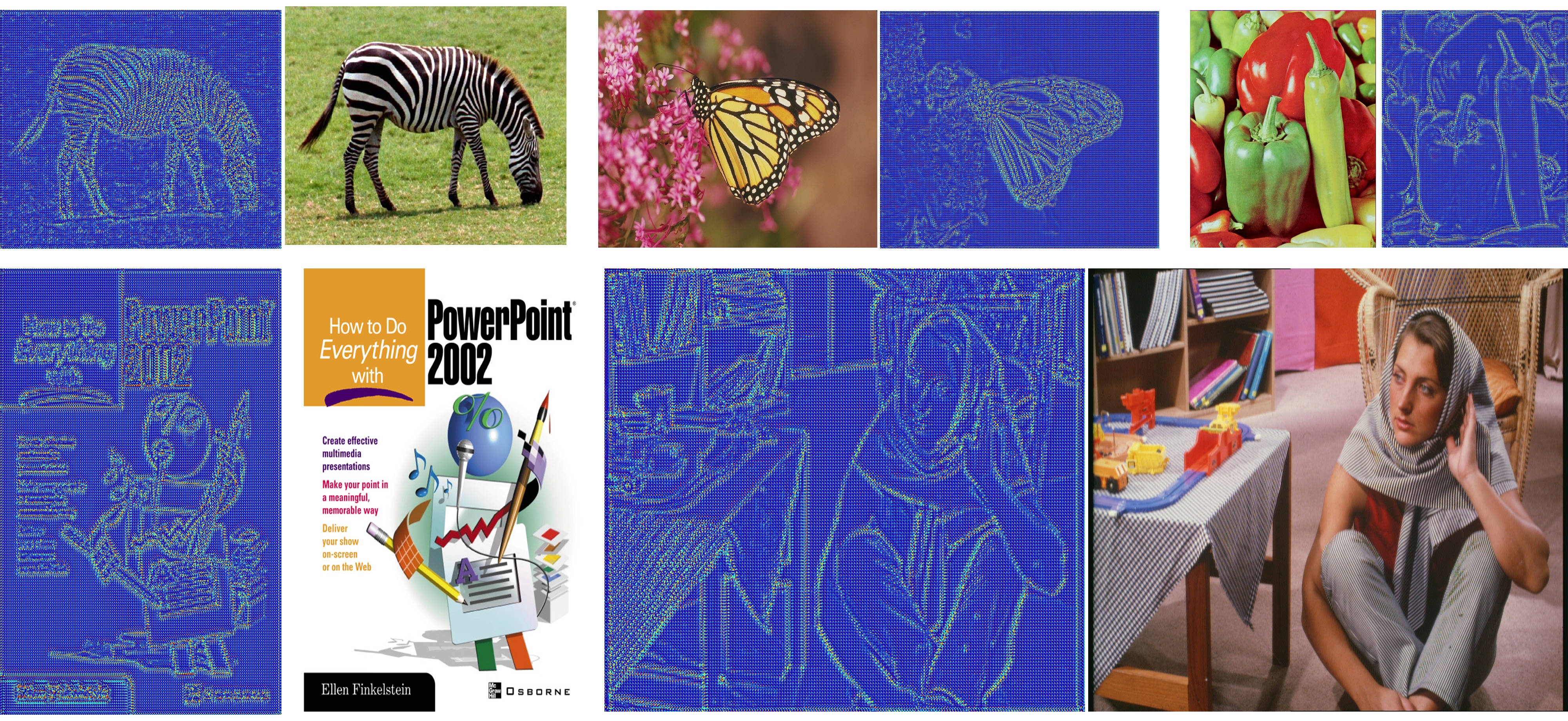}
\caption{Visualization of the position distribution.}
  \label{fig:offset}
\end{figure*}
\section{Additional Ablation Study}
\label{sec:addtional_ab_study}
Due to space limitations in the main text, we provide additional ablation experiments to demonstrate the effectiveness and rationality of the proposed method. Below, we present detailed descriptions of additional ablation studies and implementation details.\\
{\bf Ablation Study on $\mathcal{K}$.} In the DGP-Driven Covariance Weighting, considering the difficulty for deep learning networks to directly interpret the specific meaning of covariance, we map $\mathcal{K}$ from its original three-dimensional representation (\textit{i.e.}, $(\sigma_x^2$, $\sigma_y^2$, and $\rho$)) to a latent representation space through a convolutional neural network. This approach facilitates better convergence and achieves improved performance. Specifically, we explore the performance when the dimension of the latent space is set to 3, 256, and 512, as shown in Table~\ref{tab:ab_dim}. It can be observed that the best performance is achieved when the dimension is set to 512, showing significant improvements compared to the original three-dimensional setting. \\
{\bf Ablation Study on number of $\mathcal{K}$.} Furthermore, we investigate the impact of the number of $\mathcal{K}$ on the network's performance. We define 100, 500, and 730 Gaussian covariances, and the results are presented in Table~\ref{tab:ab_k}. It can be seen that as the number of covariances increases, the network's performance improves. However, beyond 730, no further performance gains are observed. Therefore, in this work, we set the number of covariances to 730. \\
{\bf Ablation Study on $N$.} Additionally, we study the effect of the number of Gaussians $N$ on the network's performance. As described in the main text, we initialize one Gaussian kernel at the center of each LR pixel. We further explore the impact of introducing more Gaussian kernels per unit pixel, and the results are shown in Table~\ref{tab:ab_n}. It can be observed that the best performance is achieved when the number of kernels is set to 4. Introducing too many kernels increases the optimization complexity, which does not lead to further performance gains. Therefore, we set the number of kernels per pixel to 4 in this work.\\
{\bf Ablation Study on $P_\text{off}$.} Finally, we examine the impact of the range of $P_\text{off}$, denoted as $[-A, A]$, on the network's performance. Specifically, we set the range to 0.5, 1, and 2, and the results are shown in Table~\ref{tab:ab_off}. It can be observed that the best performance is achieved when the range is set to 1. Both overly large and overly small ranges have adverse effects on performance.

\begin{table}[t]
\centering

\caption{Comparison with GaussianImage~\cite{zhang2024gaussianimage}.}
\scalebox{0.90}{
\begin{tabular}{l|cccccc}
\toprule
  & LIIF  & LTE & CiaoSR & GaussianImage & Ours \\ 
\midrule
Set5     & 32.73& 32.81 & 32.84 & 28.01         & \textbf{33.24} \\
Set14    & 28.98  & 29.06 & 29.08 & 25.62         & \textbf{29.40} \\
\bottomrule
\end{tabular}
}
\label{tab:GaussianImage}
\end{table}

\section{Additional Visual Comparison Results}
\label{sec:addtional_visual_comp}
\begin{figure*}[h]
  \centering
  \includegraphics[width=1\linewidth]{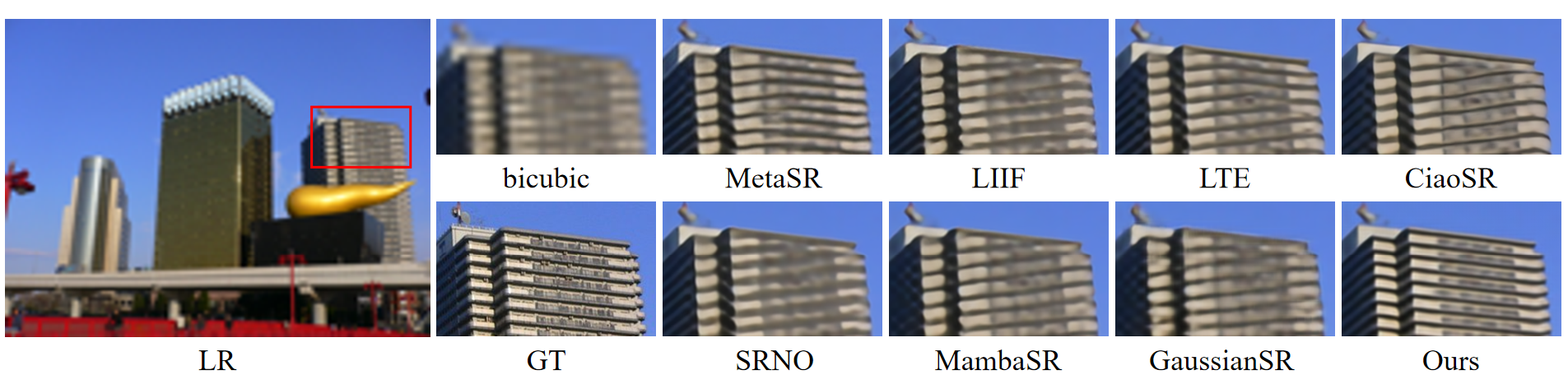}
\caption{More qualitative comparison. The visual quality of our method outperforms existing methods. Please zoom in for a better view.}
  \label{fig:vis_more1}
\end{figure*}

\begin{figure*}[h]
  \centering
  \includegraphics[width=1\linewidth]{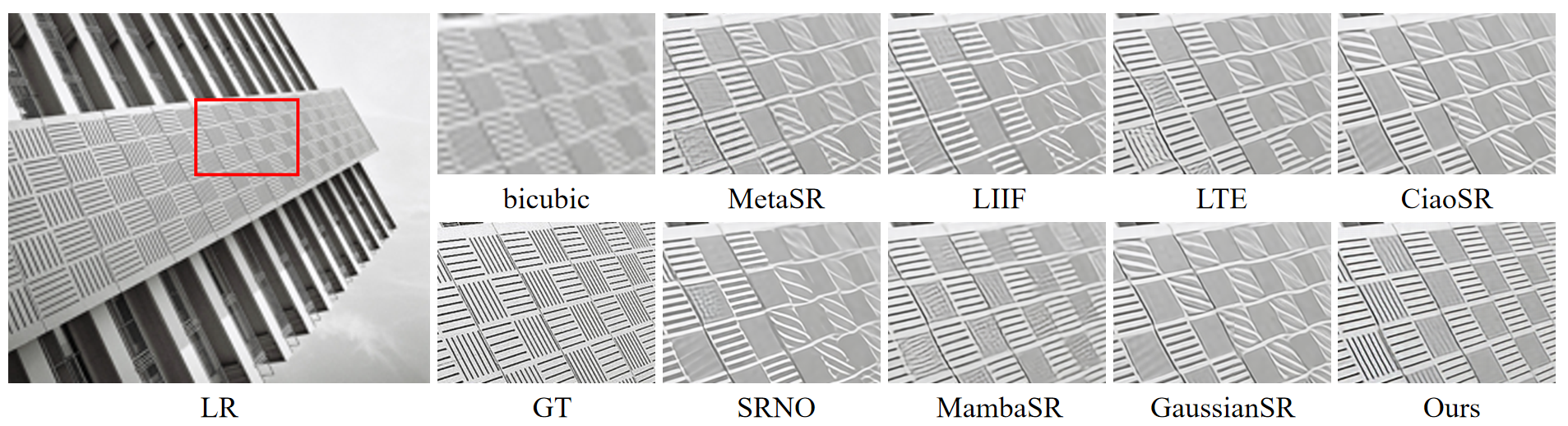}
\caption{More qualitative comparison. The visual quality of our method outperforms existing methods. Please zoom in for a better view.}
  \label{fig:vis_more2}
\end{figure*}

\begin{figure*}[h]
  \centering
  \includegraphics[width=1\linewidth]{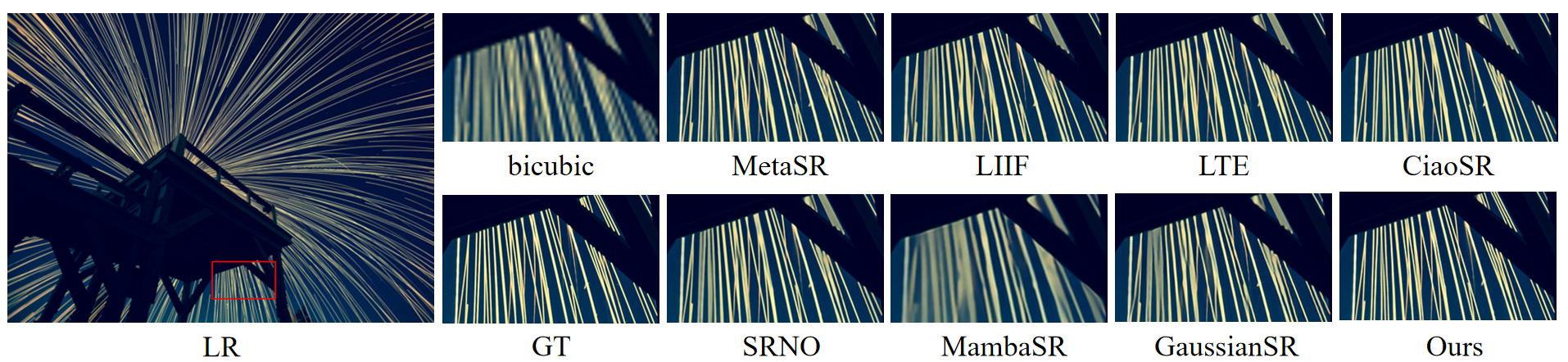}
\caption{More qualitative comparison. The visual quality of our method outperforms existing methods. Please zoom in for a better view.}
  \label{fig:vis_more3}
\end{figure*}

In this section, we present additional visual comparison results to further demonstrate the superiority of our proposed method, as shown in Figure ~\ref{fig:vis_more1},~\ref{fig:vis_more2} and~\ref{fig:vis_more3}. It can be observed that our method achieves the best visual satisfaction in terms of detailed textures, while also preserving the highest level of detail fidelity, making it closest to the GT image.

\section{Visualization of the Position Distribution}
\label{sec:offset}

To demonstrate the superior adaptive perception capability of the proposed offset mechanism, which effectively introduces more Gaussian kernels in complex texture regions based on image content, we visualize the learned Gaussian position distribution. As shown in Figure~\ref{fig:offset}, the proposed Adaptive Position Drifting adjusts the original initialization of the position distribution by adaptively perceiving the structural content of the image. The results reveal that regions with richer textures have higher densities of Gaussian kernels.

\section{Algorithm Workflow}
\label{sec:algorithm}

To clearly demonstrate the details of the proposed method, we design an algorithm workflow, as illustrated in Algorithm \ref{alg:gaussian_splatting}. This workflow describes the key steps from input to output, including feature encoding, color prediction, offset prediction, covariance estimation, and the final image reconstruction process.

\section{More Exploration and Results}
In Section 6 of the main text, we demonstrate the performance of our method in low-resolution and rainy scenarios. Here, we present comparisons across more scaling factors and with the two-stage ASSR+Derain methods DRSformer~\cite{Chen_2023_CVPR}. The results are shown in Table~\ref{tab:ASSR_Derain2}. As shown in Table~\ref{tab:ASSR_Derain2}, our method consistently outperforms other methods across all scaling factors. For instance, at the $\times 4$ scale, our method achieves a PSNR of 24.51, significantly higher than the best two-stage method, DRSformer~\cite{Chen_2023_CVPR}+LIIF (20.14). At the $\times 8$ scale, our method achieves 22.76, outperforming DRSformer+LIIF (18.93). Compared to other "All in one" methods, our approach also achieves superior results, such as 23.95 at the $\times 5$ scale, outperforming both GaussianSR (23.51) and CiaoSR (23.45). These results highlight the robustness, simplicity, and effectiveness of our method for super-resolution and deraining tasks across various scales.

\RestyleAlgo{ruled}
\SetKwInput{KwInput}{Input}
\SetKwInput{KwOutput}{Output}
\begin{algorithm}[t]
\caption{\NAME.} 
\KwInput{$\text{inp} : \text{(B, 3, H}_0\text{, W}_0\text{)}$} 
\KwOutput{$\text{image} : \text{(B, 3, H, W)}$} 

$F \gets \text{Encoder}(\text{inp})$\;
$C \gets \text{CGM}(F)$ \tcp*[l]{$[N \times 3]$}
$\Delta x \gets \text{APD}(F)$ \tcp*[l]{$[N \times 2]$}
$\Sigma \gets \text{DDCW}(F)$ \tcp*[l]{$[N \times 3]$}

$\tilde{x} \gets \text{Grid}(H, W) + \alpha \Delta x$ 

\For{$n \in [1, N]$}{
    $(xys_n, \text{depth}_n, \text{radii}_n, \text{conic}_n) \gets \text{Project}(\tilde{x}_n, \Sigma_n)$\;
}

$\text{image} \gets \text{Composite}(\{xys, \text{depth}, \text{radii}, \text{conic}, C\})$\;
\label{alg:gaussian_splatting}  

\end{algorithm}

\section{Limitation and Future Work}
{\bf Position Distribution.} In this paper, to address the difficulty of optimizing position parameters, we propose \NAMEPos, which leverages an offset mechanism to alleviate the optimization challenges and enhance the representational capacity of the model. However, assigning one or four Gaussian kernels to each LR pixel introduces some limitations. On the one hand, it leads to an overabundance of Gaussian kernels in low-frequency regions, resulting in resource wastage. On the other hand, it increases the optimization difficulty significantly. To address these issues, we plan to explore the adaptive allocation of Gaussian kernels based on the texture complexity of image content in future work. This approach aims to dynamically assign an appropriate number of kernels to different regions, effectively mitigating the aforementioned problems. \\
{\bf Introduce Generation Knowledge.} In addition, considering that arbitrary-scale super-resolution sometimes requires large upscaling factors (\textit{e.g.}, ×16, ×32, \textit{etc.}), it is challenging for the model to generate high-quality details solely relying on the input image and model knowledge. Therefore, in the future, we plan to incorporate more visual knowledge from diffusion models or semantic knowledge from large vision-language models to help the network generate finer details for high-magnification scenarios.

\end{document}